# Multivariate time series classification with dual attention network

Mojtaba A. Farahani
IMSE department
West Virginia University
Morgantown, 26505, WV, USA
ma00048@mix.wvu.edu

Tara Eslaminokandeh
MIE department
Northeastern University
Boston, 02115, MA, USA
eslaminokandeh.t@northeastern.edu

*Abstract*— One of the topics in machine learning that is becoming more and more relevant is multivariate time series classification. Current techniques concentrate on identifying the local important sequence segments or establishing the global long-range dependencies. They frequently disregard the merged data from both global and local features, though. Using dual attention, we explore a novel network (DA-Net) in this research to extract local and global features for multivariate time series classification. The two distinct layers that make up DA-Net are the Squeeze-Excitation Window Attention (SEWA) layer and the Sparse Self-Attention within Windows (SSAW) layer. DA-Net can mine essential local sequence fragments that are necessary for establishing global long-range dependencies based on the two expanded layers.

## I. INTRODUCTION

Time-series data is one of the main data types that are available in all domains. It is becoming more ubiquitous thanks to the increasing number of sensors and sensing technologies and can be found across a wide range of industries. Stock market prices in financial markets, ECG data in healthcare[1], Ecohydrology sensing data[2], positional data from our smart wearable devices, high-resolution images from the sun over time[3], 3D depth sensor Kinect data[4], and vibration, pressure, and temperature data coming from manufacturing sensors are all examples of time-series data. Time-series classification (TSC) is considered one of the most challenging problems in data mining that has many real-world applications (e.g., quality inspection[5], fault detection in smart manufacturing systems[6], demand forecasting in power systems[7], mechanical behavior prediction[8], etc.). In particular, time series classification can be categorized into Univariate Time Series Classification (UTSC) and Multivariate Time Series Classification (MTSC). Data generated by a single sensor are referred to as univariate time series, and data generated simultaneously by multiple sensors are referred to as multivariate time series. There is an increasing demand for the development of MTSC models for real-life applications. To this end, this paper review the newly proposed MTSC algorithm titled DA-NET[9] in detail and them we implement the algorithm on benchmark datasets. We also compared the results with few other algorithms.

The motivation behind this work is the fact that besides focusing on algorithm development and analysis, it is essential to concurrently undertake the validation and empirical comparison of the numerous existing algorithms. This endeavor holds immense value for practitioners as it narrows their options and provides insights into the strengths and weaknesses of available models. Additionally, this effort helps to direct research towards more promising avenues.

## II. RELATED WORKS

A vast variety of MTSC approaches have been developed in the literature. They can be divided into traditional machine learning-based methods and deep learning-based methods. Here, we briefly describe and introduce them.

### A. Machine learning-based methods

MTSC is a well-stablished topic in the machine learning community. Various approaches have been proposed to do this task. We categorize conventional ML techniques from two separate points of view (i.e., Feature engineering technique and Classification technique). We summarize each in several groups as follows:

**Feature engineering techniques**: Distance-based techniques (e.g., in Fulcher and Jones's feature-based linear classifier (FBL) algorithm[10]), dictionary-based techniques (e.g., in Bag of Patterns (BOP)[11], Symbolic Aggregate Approximation Vector Space Model (SAXVSM)[11], Bag of SFA Symbols (BOSS)[12], and WEASEL+MUSE[13]), differential-based techniques (e.g., in KNN-CID, KNN-DDTW, KNN-DTDC algorithms[11]), image-based techniques (e.g., in Recurrence Plot Compression Distance (RPCD) algorithm[11]), interval-based techniques (e.g., in Time Series Forest (TSF) algorithm[14], Random Interval Spectral Ensemble (RISE) algorithm[15], and Canonical interval forest (CIF) algorithm[16]), shapelet-based techniques (e.g., in Learned Shapelets (LS), Fast Shapelets (FS), and Shapelet Transform (ST) algorithms[11]), hybrid techniques (e.g., in TS-CHIEF[17] and HIVE-COTE V2.0[18] algorithms ), and statistical techniques (e.g., in rotation forest algorithm[19]) are among the most used feature engineering techniques in the literature. The detail of each technique is out of the scope of this report. There are also many algorithms that don't do any kind of feature engineering on the data and do the classification directly on the raw time-series (e.g., in KNN-DTW and KNN-LCSS algorithms[11], and XGBoost[20]).

**Classification Techniques:** From classification point of view, Statistical classifiers (e.g., in Learned Shapelets (LS)[11], The multiple representation sequence learner (MrSEQL)[21], The random convolutional kernel transform (ROCKET)[22] algorithms), Distance-based classifiers (e.g., in KNN-DTW, KNN-ED[11], and SVM algorithms), Algorithm Ensemble classifiers (e.g., in Elastic Ensemble (EE)[11] and Contractable Bag of Symbolic Fourier Approximation Symbols (CBOSS)[23] algorithms), Decision tree Ensemble classifiers (e.g., in Proximity Forest (PF)[24], Canonical interval forest (CIF)[16], and TS-CHIEF[17] algorithms ), and Meta Ensemble classifiers (e.g., in HIVE-COTE V1.0 and HIVE-COTE V2.0[18] algorithms) are the main classification techniques in the literature. Researchers use on raw time-series data or combine with previously mentioned feature engineering techniques to do the UTSC and MTSC tasks. Again, the details of these techniques are out of the scope of this report.

*B. Deep learning-based methods*

Although conventional time-series classification algorithms are sometimes very powerful and are capable of generate very accurate results, they lack scalability when dealing with large datasets and their results depend on domain knowledge and hand-crafted features. Thus, researcher's attention has been shifted toward deep leaning-based methods to deal with these shortcomings. Its biggest advantage is that deep learning can extract features with the help of neural networks. Among these methods of neural networks, the most representative approach is the CNNs.

CNNs have shown good results in time series classification due to their powerful local feature capture capabilities. OS-CNN[25], InceptionTime[26], and RTFN[27] are among CNN-based algorithms. They aim to find the local patterns by high-dimensional nonlinear feature extraction. However, long-range dependencies are often not considered in these approaches.

Several studies on the long-range dependency information have been proposed. Tscaps[28], Att-dGRU-SE-FCN[29], and TCN[30] are examples of these techniques. Besides this, researchers investigate a new flood of artificial intelligence in addition to convolutional-based neural networks Transformer. For example, BERT and GPT2[31], have gained prominence through their remarkable results in the Nature Language Processing (NLP) due to the effective self-attention different kinds of time-series classification applications.

Recently published findings have shown that modeling the local–global features is essential. For example, CTNet[32], LMP[33], MLSTM-FCN[34] are examples of these methods.

Although existing methods perform well, they fail to handle the long sequence with the local discriminate subsequences adequately. To address this issue, the authors of this paper propose dual attention to discover the local patterns and the global information and for MTSC.

## III. PROPOSED METHODOLOGY

We review the definition of MTSC in this part first. Then, we give a thorough explanation of DA-Net, including its network topology and all proposed mechanism's workflow.

*A. Problem Description*

We define $X = \{X_1, X_2, \ldots, X_N\}$ a set of multivariate time series of N instances, where each instance $X_i$ is a multivariate time series with T timestamps $t \in [1, T]$ and C dimensions $c \in [1, C]$ (the multivariate time series data are referred to situations with c exceeding 1; the univariate time series data are referred to situations with c equal to 1). We define a mapping $y = f^*(x; \theta)$ where the objective of classification model is to approximate function f, by learning the nonlinear embedding parameters h of model.

*B. Hierarchical structure*

Fig. 1 illustrates the DA-Net's overall structure. The structure consists of dual-attention blocks and four time-block partition layers. The goal of time-block partition layers is to create a hierarchical structure and shorten time series. Through the building of window-window relationships and the use of sparse attention, dual-attention blocks aim to learn local-global properties.

The time-block partition layer is executed at the beginning of each stage. Here we denote the time series of the first stage as $X \in \mathbb{R}^{T \times C}$. The time-block partition layer first concatenates 4 non-overlap neighbor timestamps as a time-block, which is analogous to tokens in NLP, thus obtaining $\frac{T}{4}$ time-blocks. Then each time-block is flattened and projected to a 4C dimensional embedding. Finally, the time series data $X \in \mathbb{R}^{\frac{T}{4} \times 4C}$ are fed into the dual-attention block, where the dimensions of time series keep the same after the block. In consecutive stages 2, 3 and 4, the process of stage 1 is repeated.

The dual-attention block can be split into 2 consecutive modules. Fig. 2 shows an illustration of a series of essential layers inside each module. The first module consists of a SEWA layer, a SSAW layer, a Layer Normalization (LN) layer, and a MLP layer. The only difference between the first and second modules is the introduction of shifted window layer[35], which shifts time-blocks within the window to resolve the problem that long-time dependencies are restricted to local window partitioning.

*C. SEWA Layer*

The SEWA layer, inspired by the success of SeNet[29], is applied to the Transformer that enables the network to focus on the local distinguishing features with the aid of global window-features. The SEWA layer divides features by

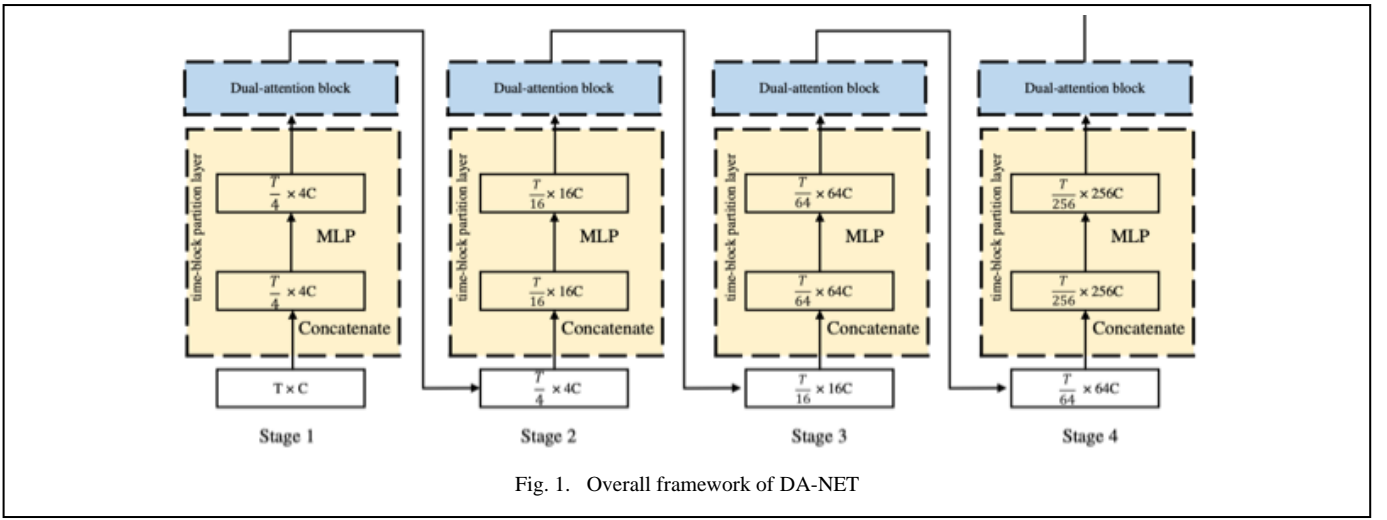

Fig. 1. Overall framework of DA-NET

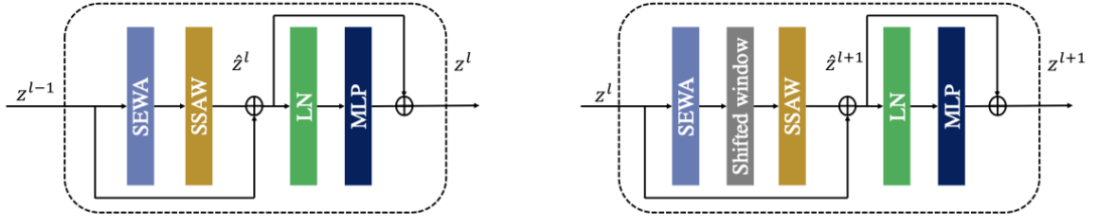

(a) The first module of dual-attention block    (b) The second module of dual-attention block

Fig. 2. A whole dual-attention block involves two components, where $\hat{z}_l$ and $z_l$ denote the output features of SSAW layer and MLP layer for block l, respectively

windows into $X \in \mathbb{R}^{M \times C \times W}$, where M stands for the number of windows, C stands for the number of channels, and W stands for the number of the time block within a window. The SEWA layer is a computational unit building on a feature transformation mapping an input $X \in \mathbb{R}^{M \times C \times W}$ to feature $S \in \mathbb{R}^{M' \times C' \times W'}$. The two-step process of SEWA layer contains squeeze and excitation operations, as shown in Fig. 3.

**Squeeze:** This step aims to get information between windows based on long-range dependency relationships. We average all time-blocks of the windows to obtain the weights of each window. More specifically, we use global average pooling to aggregate window information and generate a window-wise contextual descriptor for each window. The squeeze operation is defined as follows:

$$Z = F_{sq}(X) = \frac{1}{C \times W} \sum_C \sum_W X(c, w) \quad (1)$$

**Excitation:** Current descriptors do not have contextual window information and they cannot establish local distinguishing features with global information. To address this issue, we follow squeeze operation with two linear projections to learn the weight of contextual window information:

$$H = F_{ex}(Z) = W_2 . ReLU(W_1 Z) \quad (2)$$

where W1 and W2 stand for the learning parameters of linear projections.

Finally, the SEWA layer applies the window-wise contextual descriptors to the original features and thus suppressing non-significant windows and amplifying significant ones. We use the sigmoid function to scale the weights varying from 0 to 1. The normalized weights are weighted to the original features by a final multiplication operation on each window. The equation can be defined as follows:

$$S = F_{scale}(H, X) = X . Sigmoid(H) \quad (3)$$

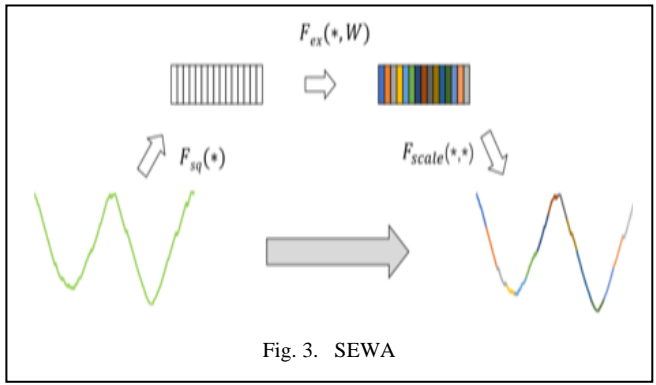

Fig. 3. SEWA

It is capable of implicitly embedding the high-level features of multivariate time series into the windows, as the SEWA layer sums the values from time-blocks and channels.

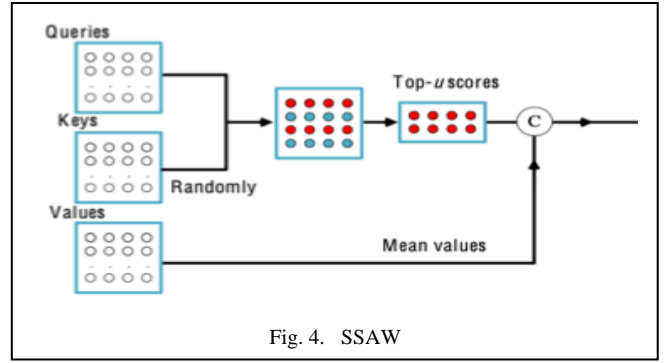

Fig. 4. SSAW

### D. SSAW Layer

To capture the global long-range dependencies, Multi-head attention based on the window (W-MHA) traversals all the queries in a window for calculating each dot-product pair. By increasing the window size, the receptive view of the window is expanded. The widen window can enhance the representation of sequences and improve the network classification results. However, W-MHA is subject to the limitations that does not allow the window size to increase indefinitely.

Thus, this paper proposes the SSAW layer to reduce the computation complexity and thus expand the window size. SSAW selects top-u by Kullback–Leibler divergence followed by self-attention calculation. SSAW can handle longer series than W-MHA.

Like W-MHA, SSAW accepts queries $Q \in \mathbb{R}^{M_q \times C}$ keys $K \in \mathbb{R}^{M_k \times C}$, and values $V \in \mathbb{R}^{M_v \times C}$ as inputs, where $M_q$, $M_k$ and $M_v$ stand for the window size. To be specific, the difference between W-MHA and SSAW is that SSAW reduces the scale of keys and takes top-u dominant queries (see Fig. 4), thus reducing the computation of scores. we calculate the max-mean measurement $\bar{M}$ of the i-th query according to the following formula:

$$\bar{M}(q_i, K) = Max_j \left\{ \frac{q_i K_j^T}{\sqrt{d}} \right\} - \frac{1}{L_k} \sum_{j=1}^{L_k} \frac{q_i K_j^T}{\sqrt{d}} \quad (4)$$

The formulation consists of the max-operator and mean-operator. The max-operator implies that i-th query has a notable variance on keys, while the mean-operator suggests that i-th query is closer to the uniform distribution on keys. Subtraction of the two operations aims to prioritize the top-u dominant queries Q. Here we define the top-u dominant queries Q as follows:

$$\bar{Q} = top - u\ \bar{M} \quad (5)$$

Finally, we concatenate the Q and the mean scores of Vs to obtain self-attention feature map. If the i-th query is selected, the corresponding score is calculated. Otherwise, the mean value of V is used instead of the calculation of the score value, which will greatly reduce the computational effort. The output of feature map S is shown below:

$$S = \begin{cases} Softmax\left(\frac{\bar{Q}K^T}{\sqrt{d}}\right).V & \text{if } i-th\ query\ is\ in\ top - u \\ mean\ (V) & otherwise \end{cases} \quad (5)$$

## IV. RESULTS

This section illustrates its modularity and flexibility by applying the proposed DA-Net for MTSC. Utilizing a series of well-established benchmarks, they conduct classification experiments to verify the performance of DA-Net. The source code and data of DA-Net are freely available athttps://github.com/Sample-design-alt/DANet[9].

### A. Experiment Setting

We build the network with these hyper-parameters: batch size B = 16, window size M = 64, the channel number of hidden layer C = 96, multi-head numbers = {3, 6, 12, 6}, and layer numbers = {2, 2, 6, 2}. Additionally, the optimization is performed using ADAM optimizer with parameters $\alpha = 1e - 3$, $\beta_1 = 0.9, \beta_2 = 0.999$, and $\epsilon = 1e - 8$. Minimization is performed using cross entropy loss function. We use the same strategies in all conducted experiments: optimizer, data augmentation and train/test datasets. All results are obtained with 100 iterations.

### B. Datasets Details

We collect benchmark datasets[11] for MTSC to test our methodology. This repository contains 30 datasets and is grouped in different application areas. The dimensions of these datasets range from 2 in trajectory to 1345 in the classification task. The non–homogeneous datasets are packed with different time series lengths from 8 to 17984 and large sizes from 27 to 50000. The list of 30 datasets, and their characteristics are illustrated in Fig. 5.

### C. Evaluation Metrics

In this paper, we evaluate the proposed methodologies using ''AVG acc'', ''Win'', ''AVG rank'', and ''MPCE'' indexes[36]. Note that ''AVG acc'' and ''Win'' indexes denote the average accuracy of methodology and the number of wins on multiple datasets, respectively. ''AVG rank'' index is defined to measure the corresponding and best methodology difference. ''MPCE'' index is a valid method that calculates the mean error rates by taking the factor of category amount into account. The equation is defined as follows:

$$MPCE = \frac{1}{k} \sum_{k=1}^{k} \frac{e_k}{d_k} \quad (6)$$

where K is the number of datasets, Dk represents the number of categories in the k-th dataset, and ek represents the error rate on the k-th dataset. Smaller ''AVG rank'' and ''MPCE'' values indicate better performance of the methodology.

| Dataset | Character | | | | | | |
|---|---|---|---|---|---|---|---|
| | Abbreviation | Type | Train | Test | Dimensions | Length | Classes |
| ArticularyWordRecognition | AWR | Motion | 275 | 300 | 9 | 144 | 25 |
| AtrialFibrillation | AF | ECG | 15 | 15 | 2 | 640 | 3 |
| BasicMotions | BM | HAR | 40 | 40 | 6 | 100 | 4 |
| CharacterTrajectories | CT | Motion | 1422 | 1436 | 3 | 182 | 20 |
| Cricket | CR | HAR | 108 | 72 | 6 | 1197 | 12 |
| DuckDuckGeese | DDG | AS | 50 | 50 | 1345 | 270 | 5 |
| EigenWorms | EW | Motion | 128 | 131 | 6 | 17984 | 5 |
| Epilepsy | EP | HAR | 137 | 138 | 3 | 206 | 4 |
| EthanolConcentration | EC | HAR | 261 | 263 | 3 | 1751 | 4 |
| ERing | ER | Other | 30 | 270 | 4 | 65 | 6 |
| FaceDetection | FD | EEG/MEG | 5890 | 3524 | 144 | 62 | 2 |
| FingerMovements | FM | EEG/MEG | 316 | 100 | 28 | 50 | 2 |
| HandMovementDirection | HMD | EEG/MEG | 160 | 74 | 10 | 400 | 4 |
| Handwriting | HW | HAR | 150 | 850 | 3 | 152 | 26 |
| Heartbeat | HB | AS | 204 | 205 | 61 | 405 | 2 |
| InsectWingbeat | IW | AS | 30000 | 20000 | 200 | 30 | 10 |
| JapaneseVowels | JV | AS | 270 | 370 | 12 | 29 | 9 |
| Libras | LIB | HAR | 180 | 180 | 2 | 45 | 15 |
| LSST | LSST | Other | 2459 | 2466 | 6 | 36 | 14 |
| MotorImagery | MI | EEG/MEG | 278 | 100 | 64 | 3000 | 2 |
| NATOPS | NA | HAR | 180 | 180 | 24 | 51 | 6 |
| PenDigits | PD | Motion | 7494 | 3498 | 2 | 8 | 10 |
| PEMS-SF | PEMS | Other | 267 | 173 | 963 | 144 | 7 |
| Phoneme | PM | AS | 3315 | 3353 | 11 | 217 | 39 |
| RacketSports | RS | HAR | 151 | 152 | 6 | 30 | 4 |
| SelfRegulationSCP1 | SRS1 | EEG/MEG | 268 | 293 | 6 | 896 | 2 |
| SelfRegulationSCP2 | SRS2 | EEG/MEG | 200 | 180 | 7 | 1152 | 2 |
| SpokenArabicDigits | SAD | AS | 6599 | 2199 | 13 | 93 | 10 |
| StandWalkJump | SWJ | ECG | 12 | 15 | 4 | 2500 | 3 |
| UWaveGestureLibrary | UW | HAR | 120 | 320 | 3 | 315 | 8 |

| Dataset | Methodology | | | | | |
|---|---|---|---|---|---|---|
| | ED-1NN | DTW-1NN-D | ED-1NN (norm) | MLSTM-FCN | DTW-1NN-I | DTW-1NN-I (norm) |
| AWR | 0.970 | **0.987** | 0.970 | 0.973 | 0.980 | 0.980 |
| AF | **0.267** | 0.200 | 0.247 | **0.267** | **0.267** | **0.267** |
| BM | 0.675 | 0.975 | 0.676 | 0.950 | **1.000** | **1.000** |
| CT | 0.964 | **0.990** | 0.964 | 0.985 | 0.969 | 0.969 |
| FD | 0.519 | 0.529 | 0.519 | **0.545** | 0.513 | 0.500 |
| HMD | 0.279 | 0.231 | 0.278 | **0.365** | 0.306 | 0.303 |
| HB | 0.620 | **0.717** | 0.619 | 0.663 | 0.659 | 0.658 |
| MI | **0.510** | 0.500 | **0.510** | **0.510** | 0.390 | N/A |
| NATO | 0.860 | 0.883 | 0.850 | **0.889** | 0.850 | 0.850 |
| PEMS | 0.705 | 0.711 | 0.705 | 0.699 | **0.734** | **0.734** |
| PD | 0.973 | 0.977 | 0.973 | **0.978** | 0.939 | 0.939 |
| SRS2 | 0.483 | **0.539** | 0.483 | 0.472 | 0.533 | 0.533 |
| SAD | 0.967 | 0.963 | 0.967 | **0.990** | 0.960 | 0.959 |
| SWJ | 0.200 | 0.200 | 0.200 | 0.067 | **0.333** | **0.333** |
| AVG acc | 0.642(±0.278) | 0.672(±0.307) | 0.64(±0.280) | 0.668(±0.306) | 0.674(±0.280) | 0.645(±0.326) |
| Win | 0 | 0 | 0 | 0 | 1 | 1 |
| AVG rank | 8.667 | 6.400 | 9.133 | 6.667 | 7.267 | 8.000 |
| MPCE | 1.780 | 1.657 | 1.789 | 1.698 | 1.668 | 1.870 |
| Dataset | Methodology | | | | | |
| | DTW-1NN-D(norm) | WEASAL + MUSE | TapNet | MR-PETSC | SMATE | DA-Net |
| AWR | 0.987 | 0.990 | 0.987 | **0.997** | 0.993 | 0.980 |
| AF | 0.220 | 0.333 | 0.333 | 0.400 | 0.133 | **0.414** |
| BM | 0.975 | **1.000** | **1.000** | **1.000** | **1.000** | 0.925 |
| CT | 0.989 | 0.990 | 0.997 | 0.984 | 0.987 | **0.998** |
| FD | 0.529 | 0.545 | 0.556 | 0.574 | 0.563 | **0.645** |
| HMD | 0.231 | 0.365 | 0.378 | 0.365 | **0.527** | 0.347 |
| HB | 0.717 | 0.727 | **0.751** | 0.702 | 0.727 | 0.626 |
| MI | 0.500 | 0.500 | **0.590** | 0.490 | **0.590** | 0.550 |
| NATO | 0.883 | 0.870 | **0.939** | 0.917 | 0.883 | 0.877 |
| PEMS | 0.711 | N/A | 0.751 | 0.861 | 0.763 | **0.867** |
| PD | 0.977 | 0.948 | 0.980 | 0.905 | 0.980 | **0.989** |
| SRS2 | 0.539 | 0.460 | 0.550 | 0.533 | 0.556 | **0.561** |
| SAD | 0.963 | 0.982 | 0.983 | 0.960 | 0.982 | **0.990** |
| SWJ | 0.200 | 0.333 | **0.400** | **0.400** | 0.200 | 0.400 |
| AVG acc | 0.673±(0.305) | 0.646(±0.326) | **0.728(±0.256)** | 0.721(±0.250) | 0.706(±0.292) | 0.724(±0.244) |
| Win | 0 | 1 | 5 | 3 | 2 | **8** |
| AVG rank | 4.600 | 4.000 | **2.133** | 3.400 | 2.867 | 2.733 |
| MPCE | 1.650 | 1.659 | 1.404 | 1.457 | 1.514 | **1.391** |

Fig. 6. Performance results

*D. Performance*

To evaluate the performance of DA-Net, the paper's authors applied DA-Net on 14 datasets out of 30 UEA datasets in Fig. 5. They compared their results with some well-performing competitors, e.g., ED- 1NN, DTW-1NN, variants of ED-1NN as well as DTW-1NN, MLSTM-FCN, WEASEL + MUSE, TapNet, MR-PETSC, and SMATE. They perform DA-Net 10 times for each dataset and return the average accuracy as the evaluation. Fig.6. shows the accuracy of all the methodologies mentioned above. And based on those results, they concluded that DA-Net results are superior and can compete with the state of the art in MTSC.

We were able to implement DA-Net based on the codes provided by the authors. We implemented DA-Net on other sets of 10 datasets (some of them are common with this paper's datasets and some are different). We compared the results with WEASEL+MUSE[13] and ROCKET[22] classifier that are considered state of the art in conventional ML algorithms that are designed for MTSC. The results are shown in TABLE 1. The experiments shows that although there were some cases that DA-Net could not outperform the SoTA, it could return an acceptable f1 score and thus, it can be considered as very good algorithm for MTSC. The experimental results show that the application of DA-Net in MTSC is practical, which provides a new framework among the convolution-free networks, making it possible to surpass the convolution networks significantly. It should also be noted that We only trained the algorithm for 50 epochs due to the time and computational power limitation. Implementing the full experiment with more algorithms is considered as the future works for this paper.

TABLE I. MY EXPERIMENT RESULT

| Dataset | Methodology | | |
|---|---|---|---|
| | *DA-Net* | *WEASEL+MUSE* | *ROCKET* |
| Cricket | 94% | **100%** | 100% |
| DuckDuckGeese | 46% | 44% | **47%** |
| EigenWorms | 47% | **90%** | 85% |
| LSST | **34%** | 31% | 29% |
| ArticularyWordRecognition | 97% | 98% | **99%** |
| BasicMotions | 94% | **100%** | 100% |

| Dataset | Methodology | | |
|---|---|---|---|
| | *DA-Net* | *WEASEL+MUSE* | *ROCKET* |
| Heartbeat | 57% | **64%** | 47% |
| Handwriting | **62%** | 36% | 49% |
| FingerMovements | 51% | **55%** | 52% |
| EthanolConcentration | 29% | **51%** | 39% |

## V. Conclusion

This paper proposed the SEWA and SSAW layers, which gather the local and global features to construct a dual-attention network (DA-Net) for MTSC. The SEWA layer collects contextual window features to mine the window-window relationships, dynamically recalibrating features and learning the local features. In addition, the SSAW layer reduces the computation complexity of within windows by Sparse-attention, which makes the network generalizable to mine the global long-range dependencies. The experiments demonstrate that DA-Net can achieve excellent results and outperform state-of-the-art approaches.

## Acknowledgment

This report is a summarization of the paper titled "DA-Net: Dual-attention network for multivariate time series classification" by Rongjun Chen et al., and it is written for the purpose of semester project report of CPE 520 course, West Virginia university. The methodology and all the figures are credited to them[9]. This version is not meant for any peer reviewed publication purposes.